\documentclass{article}
\usepackage{spconf,amsmath,graphicx,hyperref}
\usepackage{algorithmic}
\usepackage{subcaption}
\usepackage{amssymb}
\usepackage{graphicx}
\usepackage{lipsum}
\usepackage{cite}

\usepackage{float} 
\usepackage{tabularx}
\usepackage{booktabs}
\usepackage{pifont}
\usepackage{threeparttable}
\usepackage{array} 
\usepackage{multirow}
\usepackage{textcomp}
\usepackage{booktabs}
\usepackage{xcolor}

\usepackage{multirow}

\title{Unleashing Vision Transformer Potential in Image Quality Assessment via Global-Local Adaptive Interaction}
\name{
Yu Li$^{1,2,*}$, Puchao Zhou$^{1,*}$, Yachun Mi$^{1}$, 
Yanfeng Wu$^{1}$, Xiaoming Wang$^{2}$, Shaohui Liu$^{1}$\thanks{* Equal contribution.}%
}

\vspace{-5pt}
\address{
$^1$Harbin Institute of Technology, \quad $^2$Meituan
}
\begin{document}
\ninept
\maketitle
\begin{abstract}
In the field of Blind Image Quality Assessment (BIQA), accurately predicting the perceptual quality of authentically distorted images remains highly challenging due to the diverse and complex distortions present in natural environments. Although existing methods have achieved notable accuracy, their scalability is often constrained by the high cost of subjective annotation and the limited size of available datasets. Recent advances in large-scale pre-trained vision models have introduced powerful semantic and representational capabilities, yet their application to IQA tasks is hindered by substantial computational demands and suboptimal fine-tuning efficiency. To overcome these limitations, we introduce the Global-Local Interaction Adapter (GLIA), a novel framework that effectively harnesses pre-trained Vision Transformers through a dual-stream feature extraction mechanism coupled with interactive global-local fusion. By jointly retaining global semantic information and fine-grained local details, our approach delivers superior prediction accuracy and robustness while requiring significantly fewer trainable parameters. Extensive experiments on multiple benchmarks validate the effectiveness and superiority of our approach.
\end{abstract}
\section{Introduction}
\label{sec:intro}
With the rapid development of information technology and mobile internet, images have become the primary medium for information dissemination on online platforms. Across various domains—including social media, e-commerce, and several other specialized fields user demands for high-quality images are continually increasing, necessitating the development of robust Image Quality Assessment (IQA) methods. Traditional IQA approaches \cite{LIN, mittal2012no} rely on hand-crafted features and the advent of deep learning has led to the emergence of numerous learning-based IQA methods. Mainstream approaches utilize feature extractors such as CNNs\cite{DBCNN,HyperIQA,chen2024topiq} or transformers \cite{MUSIQ, Tres} to learn quality-related representations, which are subsequently decoded via regression heads or decoders to produce quality scores, achieving considerable accuracy. However, these methods typically depend on large amounts of manually annotated subjective scores, resulting in high data acquisition costs and limited dataset sizes, which restrict the training of large models.


Meanwhile, large-scale pre-trained vision models (e.g., ViT \cite{vit}, CLIP \cite{clip}) provide strong semantic priors, but full fine-tuning is often impractical due to computational cost and parameter redundancy.
To address these issues, several data-efficient IQA methods have been proposed. For example, DEIQT \cite{DEIQT} leverages ViT by fine-tuning the decoder, utilizing ViT’s pre-trained knowledge to achieve high performance with limited data. LIQE \cite{LIQE} adapts the CLIP model through multi-task fine-tuning and LoDa \cite{LoDa} integrates features from two pre-trained models by fusing multi-scale features extracted from a pre-trained CNN into ViT. However, these methods do not fully exploit the powerful priors inherent in pre-trained models, they rely on extensive parameter adjustment for knowledge transfer or inject additional model knowledge. Furthermore, due to the fixed input resolution of pre-trained models, most approaches employ repeated cropping or resizing to process input data, which can compromise crucial global semantic and local detail features in high-resolution images, both of which are essential for quality perception.
To overcome these limitations, we propose a novel adaption method called Global-Local Interaction Adapter, which maintains a low parameter count while fully leveraging the capabilities of pre-trained models. By integrating scaling and grid-based sampling \cite{FastVqa} techniques, our method exploits ViT’s superior semantic extraction abilities and utilizes a dual-stream interactive update mechanism to globally fuse locally sampled detail features, thereby repeatedly unleashing ViT’s potential in the quality perception domain. Our main contributions are summarized as follows:

1. We introduce a dual-stream semantic-detail feature extraction method that combines global semantic information with local details, mitigating the loss of perceptual features caused by resolution adaptation.

2. We design a global-local interaction fusion adapter that enables interaction between global information and local detail features in the latent space, unlocking the quality perception capabilities of pre-trained ViT.

3. Extensive experiments on multiple IQA benchmarks demonstrate that our method significantly outperforms existing approaches with substantially fewer trainable parameters, highlighting its effectiveness and generalization capability.
\begin{figure*}[ht]
    \centering
    \includegraphics[width=0.8\textwidth]{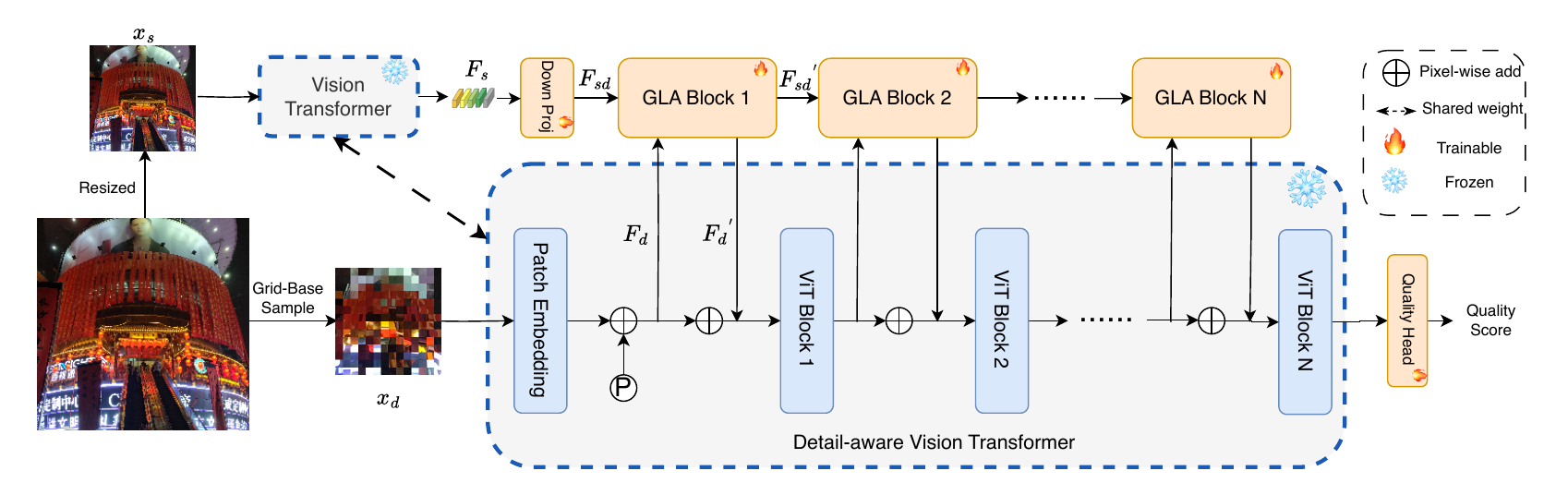}
    \caption{Framework overview of the proposed GLIANet.}
    \label{fig:backbone}
\end{figure*}
\section{The Proposed Method}
\label{sec:method}
\subsection{Overview}
\label{subsec:Arch}
As shown in Fig.~\ref{fig:backbone}, our framework, GLIANet, adopts a dual-stream architecture to preserve both global semantics and local details. The global stream resizes the input and extracts semantic tokens from a frozen ViT encoder, while the local stream samples image patches via grid-based cropping and encodes them with frozen ViT blocks. 

The two streams are fused through our Global-Local Interaction Adapter (GLIA), where semantic tokens interact with local detail features via cross-attention, enabling adaptive emphasis on distortion-aware regions. Finally, the cls token is fed into a lightweight regression head to predict the quality score. Only the GLIA, projection layers, and regression head are trainable, while the ViT backbone remains frozen.

\subsection{Semantic-Detail Dual-stream Feature Extractor}
\label{sec:GLIA}
Current IQA models often struggle to reconcile global semantics with local details, particularly when images undergo scaling or cropping.
 Inspired by recent MLLM advances \cite{zhangmllms}, which utilize cropped regions and scaled images as complementary cues, we design a dual-stream feature extractor that integrates semantic and detail representations.

The semantic stream resizes the input image and employs the pre-trained ViT to obtain global semantics $F_s$, which are further projected into a compact latent space as $F_{sd}$. The detail stream adopts a grid-based sampling strategy \cite{FastVqa}, where an image $I \in \mathbb{R}^{H \times W \times C}$ is partitioned into $n_h \times n_w$ patches of size
\begin{equation}
g_h = \left\lfloor \tfrac{H}{n_h} \right\rfloor,\quad 
g_w = \left\lfloor \tfrac{W}{n_w} \right\rfloor ,
\end{equation}
and the $(i,j)$-th fragment is extracted as
\begin{equation}
F_{i,j} = I[i \cdot g_h : i \cdot  g_h + f_h,\; j \cdot g_w : j \cdot  g_w + f_w, :],
\end{equation}
with $f_h \leq g_h, f_w \leq g_w$. Concatenated fragments form a locally magnified image that preserves fine-grained details, which are embedded by the ViT patch embedding layer to produce $F_d$.

Finally, $F_{sd}$ and $F_d$ are fused by the adapter module, enabling complementary integration of semantic priors and detail features.

\begin{figure}[ht]
    \centering
    \includegraphics[width=0.5\textwidth]{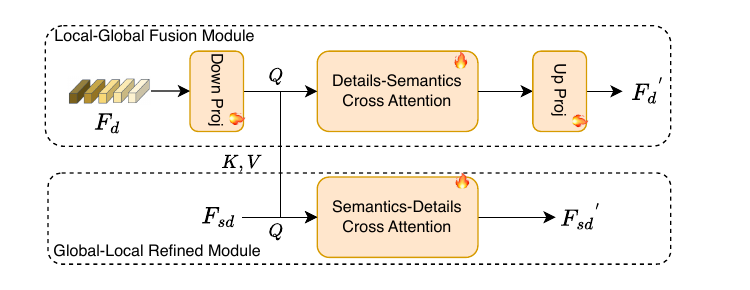}
    \caption{ The architecture of the proposed GLIA Block.}
    \label{fig:GLIA}
\end{figure}
\vspace{-10pt}

\subsection{Global-Local Interaction Adapter}
\label{sec:GLIA}

For the obtained global and local features, shown as Fig.~\ref{fig:GLIA} , we design an alignment and interaction adapter, termed the Global-Local Interaction Adapter(GLIA), which consists of two components:

\textbf{Local-Global Fusion Module}: To achieve more compact feature alignment, the local perceptual feature $F_d$ is mapped into the same latent space as $F_{sd}$ via a learnable down-mapping MLP. This step ensures that the perceptual features are compact. Subsequently, the local feature serves as the main query and the global feature as guidance; they are fused through a Multi-head Cross-Attention Mechanism(MHCA), which can be formulated as:
\begin{equation}
F^{'}_{d} = up(F_d + \lambda_d \cdot \text{MHCA}(F_d, F_{sd}))
\end{equation}
where $up$ means a up-projecting MLP and $\lambda_d$ is a learnable param for residual connection. The fused feature $F^{'}_{d}$ is then serves as the input to the next ViT-Block. This mechanism facilitates the integration of global semantic information into local details, thereby enhancing subjective image quality understanding.

\textbf{Global-Local Refined Module}: For the global semantic prior, since pre-trained models lack detailed perceptual ability, it is also updated during the interaction process. Specifically, $F_{sd}$ is used as a more refined query to interact with the latent $F_d$ via MHCA:
\begin{equation}
F_{sd}^{'} = F_{sd} + \lambda_s \cdot \text{MHCA}(F_{sd}, F_d)
\end{equation}
it should be noted that, due to the redundancy of semantic information, all operations on the semantic feature are performed in the latent space without up-projection to the original dimension. The updated prior is then used as the new global feature for interactions.

\section{Experiments}
\label{sec:exp}
\subsection{Experimental Setting}
\textbf{Datasets:} Our method is evaluated on eight classical IQA datasets, including four synthetic datasets, LIVE~\cite{live}, CSIQ\cite{CSIQ}, TID2013~\cite{tid2013},  KADID-10k~\cite{kadid10k} and four authentic datasets, LIVEC~\cite{livec}, KonIQ-10k~\cite{koniq}, SPAQ~\cite{SPAQ}, and FLIVE~\cite{flive}.

\noindent\textbf{Implementation Details:} To ensure fair comparison, we follow the same settings as LoDa \cite{LoDa}. The average Spearman Rank-Order Correlation Coefficient (SRCC) and Pearson Linear Correlation Coefficient (PLCC) are reported. All training is conducted on a single NVIDIA RTX 4090 GPU. For the pre-trained backbone, we adopt the VIT-B-16 model. Specifically, 80\% of the images are used for training and 20\% for testing, with the process repeated 10 times to minimize performance bias.

\begin{table*}[t]
   \centering
   \footnotesize
   \caption{Performance comparison measured by medians of SRCC and PLCC, \textbf{bold} entries indicate the top results and \underline{underline} indicate the second. The values in parentheses denote the number of trainable parameters during training.}
   \label{tab:allresult}
   \newcolumntype{Y}{>{\centering\arraybackslash}X}
   \begin{tabularx}{\textwidth}{@{}l *{2}{Y} *{2}{Y} *{2}{Y} *{2}{Y} || *{2}{Y} *{2}{Y} *{2}{Y} *{2}{Y}@{}}  
       \toprule
       \multirow{2}{*}{Method} & \multicolumn{2}{c}{LIVE} & \multicolumn{2}{c}{CSIQ} & \multicolumn{2}{c}{TID2013} & \multicolumn{2}{c}{KADID-10K} & \multicolumn{2}{c}{KonIQ} & \multicolumn{2}{c}{LIVEC} & \multicolumn{2}{c}{SPAQ} & \multicolumn{2}{c}{FLIVE}\\
       \cmidrule(lr){2-3} \cmidrule(lr){4-5} \cmidrule(lr){6-7} \cmidrule(lr){8-9} \cmidrule(lr){10-11} \cmidrule(lr){12-13}  \cmidrule(lr){14-15} \cmidrule(lr){16-17} 
       & SRCC & PLCC & SRCC & PLCC & SRCC & PLCC & SRCC & PLCC & SRCC & PLCC & SRCC & PLCC & SRCC & PLCC & SRCC & PLCC\\
       \midrule
       ILNIQE\cite{LIN} &0.902  & 0.906 &0.822 &0.865&0.521  &0.648  & 0.534 &0.558  &0.523  &0.537  &0.508  &0.508 &0.713  &0.712 &0.294 &0.332\\
       DBCNN\cite{DBCNN} &0.968  & 0.971 &0.946 &0.959 &0.816  &0.865  & 0.851 &0.856  &0.875  &0.884  &0.851  &0.869 &0.911  &0.915 &0.545 &0.551\\
        MetaIQA\cite{metaiqa} &0.960  & 0.959 &0.899 &0.908 &0.856  &0.868  & 0.762 &0.775  &0.887  &0.856  &0.835  &0.802 & -  & - &0.540 & 0.507 \\
        P2P-BM \cite{P2P} &0.959  & 0.958 &0.899 &0.902 &0.862  &0.856  & 0.840 &0.849  &0.872  &0.885  &0.844  &0.842 & -  & - &0.526 &0.598\\
        HyperIQA \cite{HyperIQA} &0.962  & 0.966 &0.923 &0.942 &0.840  &0.858  & 0.852 & 0.845  &0.906  &0.917  &0.859  &0.882 &0.911  &0.915  &0.544 &0.602\\
       MUSIQ \cite{MUSIQ} &0.940  &0.911  &0.871 &0.893 &0.773  & 0.815 &0.875 &0.872 &0.916  &0.928  &0.702  &0.746 &0.918  &0.921 &0.566 &0.661\\
       TReS\cite{Tres} &0.969  & 0.968 &0.922 &0.942 &0.863  &0.883  &0.859 & 0.858  &0.915  &0.928  &0.846  &0.877 & -  & -  & 0.544 & 0.625\\
       DAIQA\cite{DAIQA} & 0.969 & 0.972 & - & - & 0.863 &0.883 &0.930 & 0.925  & 0.915 & 0.925  & 0.840 &0.867 & 0.911  & 0.922  & - & - \\
       Re-IQA(\textit{48M})\cite{Reiqa} & 0.970  & 0.971 &0.945 &0.960&0.804  &0.861  &0.872  &0.885  &0.914  &0.923  &0.840  & 0.854 &0.918  &0.925 &0.575 & 0.675 \\
       QMamba(\textit{50M})\cite{Qmamba} & 0.959 & 0.958 & 0.916 & 0.935 & 0.950 & 0.952 & 0.923 & 0.938 &0.928 & 0.943  &0.863 & 0.903  &0.927  &0.933 &0.574 & 0.672  \\
       LIQE(\textit{151M})\cite{LIQE} &0.970  & 0.951 &0.943 &0.946 &-  & -  &0.930  &0.931  &0.919  &0.908  &\underline{0.904}  &0.910 & -  & - &- & -\\
       LoDa(\textit{9M})\cite{LoDa} & 0.975 & 0.979 & - & - & 0.869 &0.901 &0.931 & 0.936  & 0.932 & 0.944  & 0.876 &0.899 & 0.925  & 0.928  & 0.578 & 0.679 \\
       
       SHDIQA(\textit{24M})\cite{SHDIQA} & \underline{0.982} & \underline{0.984} & \underline{0.962} & \underline{0.973} & - & - &\underline{0.937} & \underline{0.940}  & \underline{0.937} & \underline{0.948}  & \textbf{0.909} &\textbf{0.922} & \underline{0.930}  & \underline{0.932}  & \textbf{0.639} & \textbf{0.735} \\
       \midrule
       Ours(\textit{8M}) &  \textbf{0.983} & \textbf{0.984} &\textbf{0.967}  & \textbf{0.975}  & \textbf{0.913}  & \textbf{0.923} & \textbf{0.939}  & \textbf{0.943}  & \textbf{0.940}  & \textbf{0.948}  & 0.899  & 0.913 & \textbf{0.932}  & \textbf{0.934} & \underline{0.583}  & \underline{0.688} \\
       \bottomrule
   \end{tabularx}
\end{table*}
\vspace{-5pt}
\begin{figure}[t]
    \centering
    \includegraphics[width=0.4\textwidth]{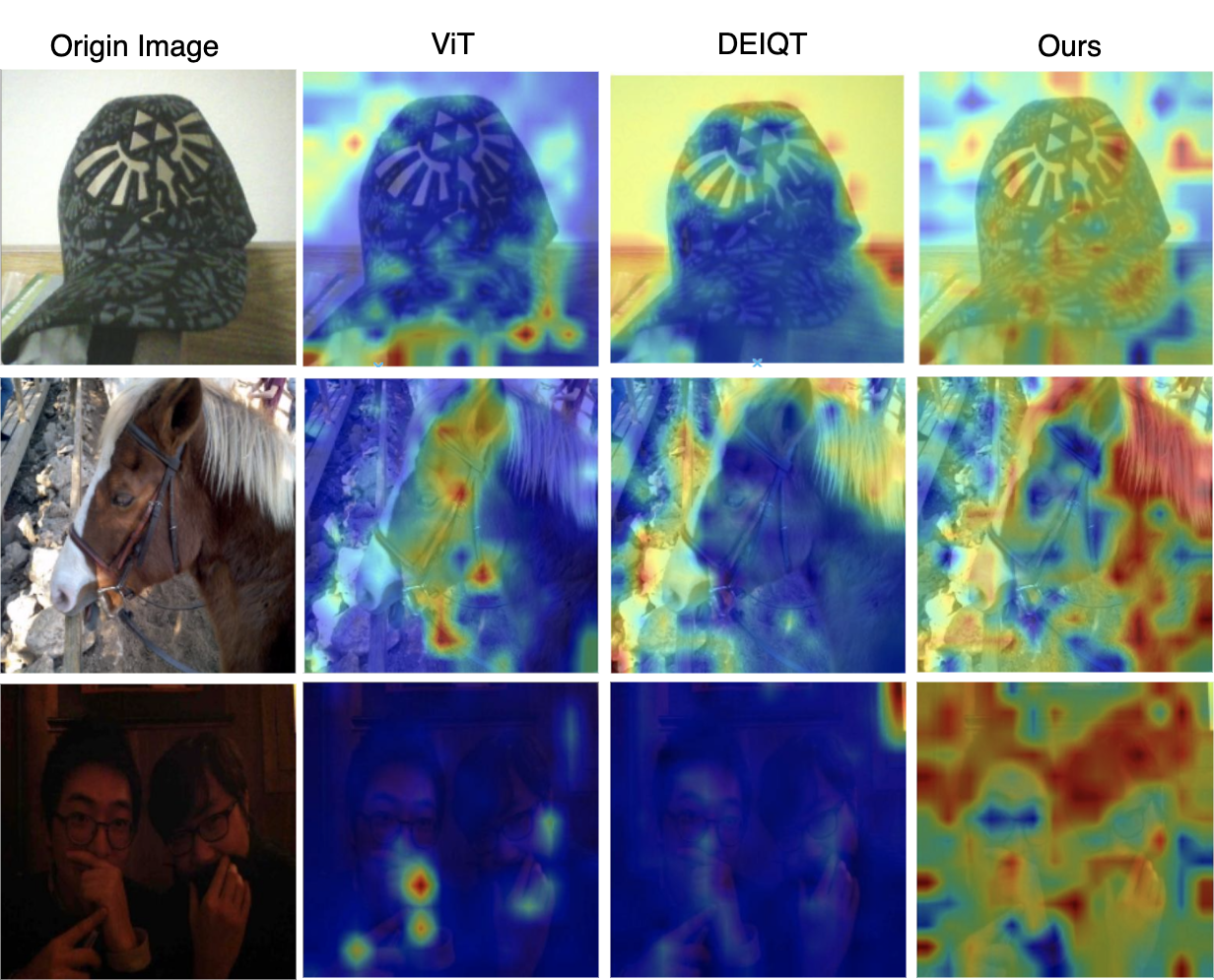}
    \caption{Visualization of the ViT attntion map.}
    \label{fig:attntion}
\end{figure}
\vspace{-5pt}
\subsection{Comparisons with the State-of-the-art Methods}
As shown as Tab.~\ref{tab:allresult} ,the comparison results between the proposed methods and other state-of-the-art (SOTA) BIQA methods. Specifically, ours surpass traditional methods (e.g., ILNIQE \cite{LIN}) and earlier learning-based methods (e.g., Tres \cite{Tres}  and HyperIQA \cite{HyperIQA} by a large margin. For LIQE \cite{LIQE} that utilized a large-scale pretrained vision-language model, multi-task labels, and full finetuning on multiple datasets simultaneously, GLIANet still outperforms on both synthetical and authentical datasets, i.e.,KADID and KonIQ. Compared with current methods that required extra pertaining (e.g., DEIQT \cite{DEIQT} , ReIQA \cite{Reiqa}  and Loda \cite{LoDa} ),our model obtains competitive or higher results, showing the effectiveness of our methods. Correspondingly, the top performance on the largest synthetical datasets KADID-10k confirms the superiority of our methods. 

\subsection{Cross-Dataset Evaluation}

To further evaluate the generalization capability of GLIANet, we follow the cross-dataset experimental protocol used in LoDa \cite{LoDa}, and additionally introduce cross-domain comparisons between two synthetic datasets. Specifically, we train the model on one dataset and test it on different datasets without any fine-tuning or parameter adjustment, to comprehensively assess the generalization ability of GLIANet against competitive BIQA models. The median SRCC results across multiple datasets are presented in Tab.~\ref{tab:Cross-Dataset}. GLIANet consistently achieves the best performance on all datasets, demonstrating its strong generalization ability. Furthermore, Fig. ~\ref{fig:attntion} illustrates the attention distributions of our GLIANet compared to DEIQT and the original ViT. Our model more precisely focuses on the subjectively important regions for quality perception, resulting in a quality assessment that better aligns with human subjective observation. 
\begin{table}[t]
   \centering
   \caption{SROCC results on Cross-Dataset evaluation.}
   \label{tab:Cross-Dataset}
   \resizebox{0.48\textwidth}{!}{
   \begin{tabular}{lcccccc}
        \toprule
        Training & \multicolumn{2}{c}{FLIVE} & LIVEC & KonIQ & LIVE & CSIQ \\
        \midrule
        Testing  & KonIQ & LIVEC & KonIQ & LIVEC & CSIQ & LIVE \\
        \midrule
        DBCNN\cite{DBCNN}   & 0.716 & 0.724 & 0.754 & 0.755 & 0.758 & 0.877 \\
        P2P-BM\cite{P2P}    & 0.755 & 0.738 & 0.740 & 0.770 & 0.712 & - \\
        TreS\cite{Tres}     & 0.713 & 0.740 & 0.733 & 0.786 & 0.761 & - \\
        DEIQT\cite{DEIQT}   & 0.733 & 0.781 & 0.744 & 0.794 & 0.781 & 0.932 \\
        LoDa\cite{LoDa}     & 0.763 & 0.805 & 0.745 & 0.811 & - & - \\
        SHDIQA\cite{SHDIQA}     & 0.723 & 0.791 & 0.782 & 0.819 & 0.789 & 0.943 \\
        \toprule
        Ours                & \textbf{0.785} & \textbf{0.816} & \textbf{0.802} & \textbf{0.835} & \textbf{0.823} & \textbf{0.953} \\
        \bottomrule
    \end{tabular}
    }
\end{table}
\begin{figure}[t]
    \centering
    \includegraphics[width=0.5\textwidth]{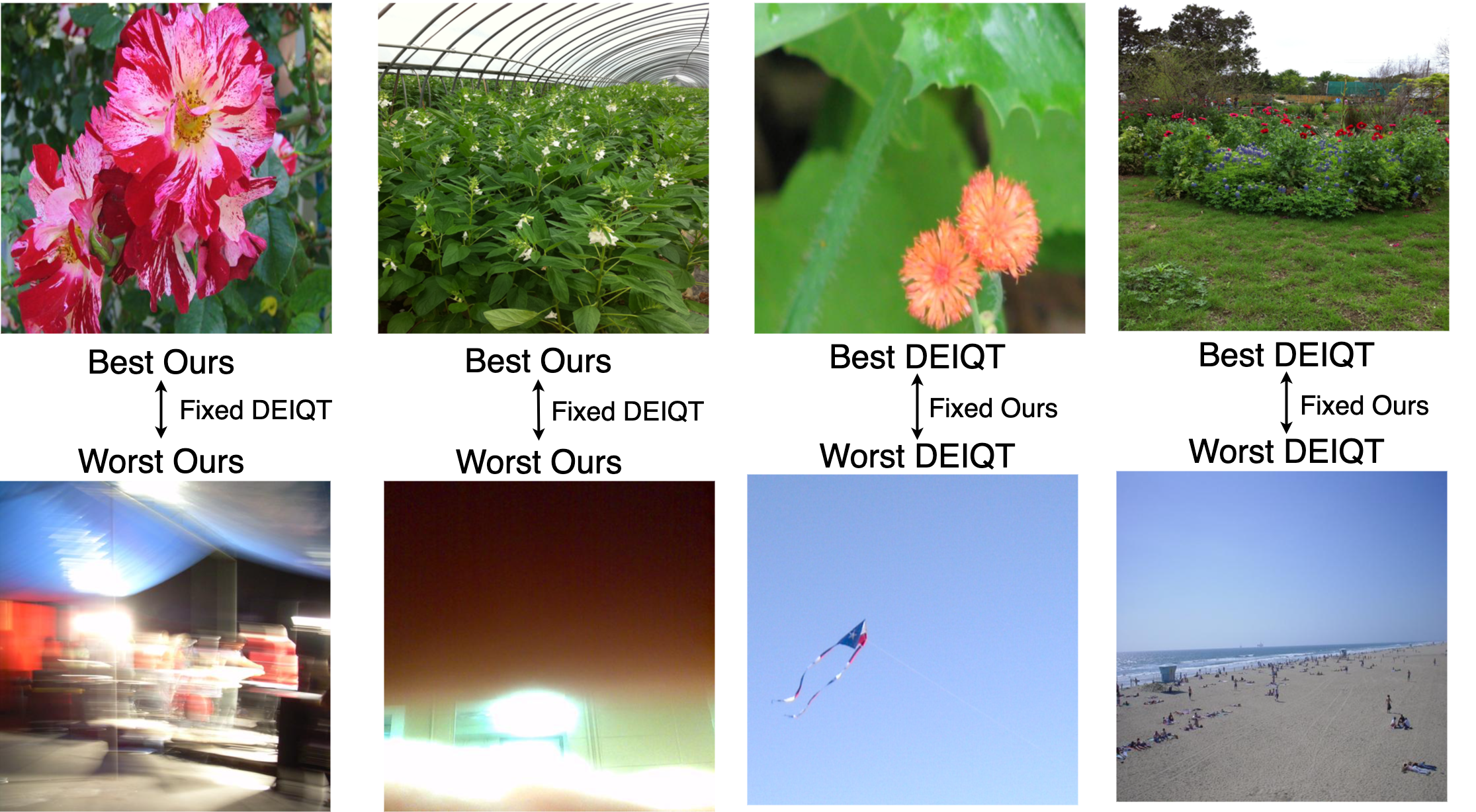}
    \caption{gMAD competition results between DEIQT and our method. From left to right: (a) Fixed DEIQT at low quality. (b) Fixed DEIQT at high quality. (c) Fixed GLIA-Net at low quality. (d) Fixed GLIA-Net at high quality.}
    \label{fig:gmad}
\end{figure}
\vspace{-9pt}
\subsection{GMad experimental model proficiency testing}
We trained the model on the Koniq-10k and tested it on the LIVEC using the gMAD competition \cite{gmad}. Image pairs are selected where the attacker model predicts a large quality difference, while the defender perceives them as similar. Observers then evaluate these pairs to assess the model’s generalization ability. As shown in Fig.~\ref{fig:gmad}, our model correctly selects image pairs with significant quality differences when attacking DEIQT, where DEIQT misclassifies pairs like (a) as low quality and (b) as high quality, contrary to human perception and GLIANet’s predictions. As a defender, GLIANet identifies pairs with minimal perceptual variation, correctly classifying image pairs like (c) and (d) as both low and high quality, respectively, while DEIQT fails. This demonstrates the generalization capability of GLIANet in handling real-world distortions.

\vspace{-9pt}
\subsection{Data-Efficient Learning Validation}

\begin{table}[t]
    \centering
    \footnotesize
    \caption{Data-efficient learning validation with the training set containing 20\%, 40\% and 60\% images.}
    \label{tab:data_efficiency}
    \begin{tabular}{cccccc}
        \toprule
         \multirow{2}{*}{Mode} & \multirow{2}{*}{Methods}  & \multicolumn{2}{c}{KonIQ} & \multicolumn{2}{c}{LIVEC} \\
        \cmidrule(lr){3-4} \cmidrule(lr){5-6}
        & & SRCC & PLCC & SRCC & PLCC \\
        \midrule
        \multirow{3}{*}{20\%}
            & HyperIQA & 0.869 & 0.873 & 0.776 & 0.809 \\
            & DEIQT    & 0.888 & 0.908 & 0.792 & 0.822 \\
            & LoDa     & 0.907 & 0.923 & 0.815 & 0.854 \\
            & Ours     & \textbf{0.917} & \textbf{0.934} & \textbf{0.843} & \textbf{0.857} \\
        \midrule
        \multirow{3}{*}{40\%}
            & HyperIQA & 0.892 & 0.908 & 0.832 & 0.849 \\
            & DEIQT    & 0.903 & 0.922 & 0.838 & 0.855 \\
            & LoDa    & 0.922 & 0.935 & 0.849 & 0.879 \\
            & Ours     & \textbf{0.932} & \textbf{0.944} & \textbf{0.873} & \textbf{0.893} \\
        \midrule
        \multirow{3}{*}{60\%}
            & HyperIQA & 0.901 & 0.914 & 0.843 & 0.862 \\
            & DEIQT    & 0.914 & 0.931 & 0.848 & 0.877 \\
            & LoDa     & 0.928 & 0.940 & 0.869 & 0.891 \\
            & Ours     & \textbf{0.935} & \textbf{0.947} & \textbf{0.890} & \textbf{0.901} \\
        \bottomrule
    \end{tabular}
\end{table}

\begin{table}[b]
\centering
\small
\caption{Ablation experiments for compositions of GLIA.}
\begin{tabular}{cccccc}
\toprule
\multirow{2}{*}{GLR} & \multirow{2}{*}{LGF} & \multicolumn{2}{c}{CSIQ} & \multicolumn{2}{c}{KonIQ} \\
\cmidrule(lr){3-4} \cmidrule(lr){5-6}
 &  & PLCC & SRCC & PLCC & SRCC  \\
 \midrule
$\checkmark$& \ding{55} & 0.951 & 0.942 & 0.939 & 0.931 \\
\ding{55}& $\checkmark$ & 0.969 & 0.961 & 0.947 & 0.937 \\
$\checkmark$& $\checkmark$ & \textbf{0.975} & \textbf{0.967} & \textbf{0.948} & \textbf{0.940} \\
\bottomrule
\end{tabular}
\label{tab:compositions}
\end{table}

\begin{table}[ht]
\centering
\small
\caption{Ablation experiments for the guidance prior.}
\begin{tabular}{c c cc cc}
\midrule
\multirow{2}{*}{Guidance} &\multirow{2}{*}{Main}& \multicolumn{2}{c}{LIVEC} & \multicolumn{2}{c}{KonIQ} \\
\cmidrule{3-6}
& & SRCC & PLCC & SRCC & PLCC \\
\midrule
$x_s$& $x_s$   & \textbf{0.901} & 0.909 & 0.932 & 0.939 \\
$x_d$ &$x_d$ & 0.876 & 0.899 & 0.930 & 0.943 \\
$x_s$ & $x_d$ & 0.899 & \textbf{0.913} & \textbf{0.940} & \textbf{0.948} \\
$x_d$  &$x_s$ & 0.883 & 0.894 & 0.936 & 0.944 \\
\midrule
\end{tabular}
\label{tab:guidance prior}
\end{table}

GLIANet  leverages the extensive prior knowledge embedded in the pre-trained ViT, enabling efficient adaptation to downstream tasks with only a small amount of data. This effectively addresses the data scarcity challenge commonly encountered in IQA, allowing our model to achieve competitive performance with others BIQA methods while significantly reducing the required training data.

Following the protocol of LoDa \cite{LoDa}, we conduct controlled experiments by training the model with limited data. The results, as shown in Tab.~\ref{tab:data_efficiency}, demonstrate that GLIANet outperforms previous models even under data-constrained scenarios. Notably, on the KonIQ-10k dataset, our model achieves competitive performance using only 60\% of the images, as indicated in Tab.~\ref{tab:allresult}.

\subsection{Ablation Study}

\textbf{Ablation Study on Dual-Stream Features :} To validate the effectiveness of our approach, we conduct ablation studies on the specific guidance priors and main features within the dual-stream architecture. The results are presented in Tab.~\ref{tab:guidance prior}. As shown, when $x_{s}$ is used as the prior, the model achieves relatively accurate performance on both datasets, attributed to the full utilization of pre-trained model knowledge. In contrast, $x_d$ does not sufficiently leverage the pre-trained priors, resulting in suboptimal performance across the datasets. The best results are obtained when the semantic feature, containing rich pre-trained prior, is used as guidance to adaptively refine the perception of local detail features, demonstrating the effectiveness of our design.

\noindent\textbf{Ablation Study on compositions of GLIA} : We further perform ablation experiments on the components of the Global-Local Interaction Adapter, as reported in Tab.~\ref{tab:compositions}. It can be observed that updating only through the Local-Global Fusion (LGF) module yields competitive results, outperforming the LoDa model. When the semantic prior is additionally updated via the Global-Local Refined(GLR) module, the performance is further improved, highlighting the importance of global-local interactive fusion.

\noindent\textbf{Ablation Study on Fine-Tuning Strategies} : Recently, various efficient model adaptation techniques have been proposed for large-scale pre-trained vision models. To validate the effectiveness of our proposed method, we conduct comparative experiments with several fine-tuning strategies: linear probing of ViT, full fine-tuning, Adapter \cite{adapter}, LoRA \cite{lora}, VPT \cite{vpt}, and LoDa \cite{LoDa} (which incorporates CNN features). These methods are evaluated on the KADID-10k and KonIQ-10k datasets, alongside our approach. The experimental results are presented in Tab.~\ref{tab:Fine-tune}. As shown, our model outperforms all these fine-tuning methods on both datasets, demonstrating the superiority of our approach.

\begin{table}[t]
\centering
\small
\caption{Ablation experiments on Fine-tune methods.}
\begin{tabular}{c cc cc}
\midrule
\multirow{2}{*}{Sample Methods} & \multicolumn{2}{c}{KADID} & \multicolumn{2}{c}{KonIQ} \\
\cmidrule{2-5}
& SRCC & PLCC & SRCC & PLCC \\
\midrule
Linear probe  & 0.677 & 0.699 & 0.794 & 0.831 \\
Full fine-tune  & 0.909 & 0.915 & 0.929 & 0.940 \\
Adapter\cite{adapter}   & 0.912 & 0.921 & 0.926 & 0.939 \\
Lora\cite{lora}  & 0.913 & 0.921 & 0.921 & 0.934 \\
VPT\cite{vpt}   &0.889 & 0.900 & 0.919 & 0.932 \\
LoDa\cite{LoDa}   &0.931 & 0.936 & 0.932 & 0.944 \\
Ours   & \textbf{0.939} & \textbf{0.943} & \textbf{0.940} & \textbf{0.948} \\
\midrule
\end{tabular}
\label{tab:Fine-tune}
\end{table}

\vspace{-9pt}
\section{Conclusion}
\label{sec:exp}
In this work, we present the Global-Local Interaction Adapter, an efficient and effective solution for image quality assessment that fully exploits the knowledge priors of pre-trained ViT. By integrating dual-stream feature extraction and global-local interactive fusion, our method overcomes the limitations of existing approaches in preserving semantic and detail information. Experimental results demonstrate that our method consistently outperforms competitive methods with reduced trainable parameters and strong generalization ability, providing a practical pathway for scalable IQA in real-world.

\clearpage
\footnotesize
\bibliographystyle{IEEEbib}
\bibliography{strings,refs}

\end{document}